\documentclass[aps,pre,superscriptaddress,notitlepage,twocolumn]{revtex4-2}

\usepackage{dcolumn}
\usepackage[T1]{fontenc}
\usepackage{amsmath, amssymb}
\usepackage{comment,xcolor,graphicx}
\usepackage{framed}
\usepackage{natbib}
\usepackage[caption=false]{subfig}

\begin{document}

\preprint{}

\title{\emph{RoboKrill}: a metachronal drag-based swimmer robot}

\author{Sara Oliveira Santos}
\affiliation{Brown University, Center for Fluid Mechanics, School of Engineering, Providence, RI, 02912, USA}

\author{Francisco Cuenca-Jiménez}
\affiliation{Universidad Nacional Autónoma de México, Engineering, Circuito Interior s/n, Coyoacán, Ciudad Universitaria, C.P. 04510, México}

\author{P. Antonio Gomez-Valdez}
\affiliation{Universidad Nacional Autónoma de México, Engineering, Circuito Interior s/n, Coyoacán, Ciudad Universitaria, C.P. 04510, México}

\author{Oscar Morales-Lopez}
\affiliation{Universidad Nacional Autónoma de México, Engineering, Circuito Interior s/n, Coyoacán, Ciudad Universitaria, C.P. 04510, México}

\author{Monica M. Wilhelmus}
\email{mmwilhelmus@brown.edu}
\affiliation{Brown University, Center for Fluid Mechanics, School of Engineering, Providence, RI, 02912, USA}


\begin{abstract}

Marine exploration is essential to understanding ocean processes and organisms. While the use of current unmanned underwater vehicles has enabled many discoveries, there are still plenty of limitations toward exploring complex environments. Bio-inspired robots are a promising solution for highly maneuverable underwater swimming at moderate speeds. Krill, especially, are efficient swimmers in the intermediate Reynolds number regime and can inform engineering solutions for ocean exploration. In this paper, we present the design, manufacture, and validation of a new krill-inspired, metachronal, drag-based robotic system. By combining active and passive actuation of the joints with 3D printed parts, our unique design recreates the swimming kinematics of \emph{Euphausia superba} in a compact and reproducible robotic platform. The motion of the anterior and posterior appendage segments is achieved using servo motors and a multi-link mechanism, while the out-of-plane motion of the biramous distal segments is attained via fluid-structure interactions. Going forward, our platform will be leveraged to study metachronal, drag-based swimmers across taxa to identify unifying success mechanisms at different scales, facilitating the development of a new generation of underwater robots.



\end{abstract}


\maketitle
\section{Introduction}
Unmanned underwater vehicles (UUV) are key for scientific marine exploration. Tethered vehicles, also known as Remotely Operated underwater Vehicles (ROV), for example, have been used to quantify the concentration of plastics in the water column at depths of up to 1 km and understand the impact of large-scale filtration by giant larvaceans \cite{Choy2019,Katija2017}. On the other hand, free-swimming Autonomous Underwater Vehicles (AUV) have been used to explore the Arctic seafloor and have facilitated the survey of marine protected areas as well as execute search and find missions \cite{Kunz2008,Benoist2019,Boulares2021}. While AUV are advantageous given their high speeds and broad spatial coverage, they are not suited for highly complex environments, such as caves, coral reefs, and shipwrecks, where the risk of collision and damage can be significant. In addition, their lack of on-board signal processing and navigation constraints result in limited maneuverability \cite{Hudson2005}. 

Bio-inspired designs have the potential to produce engineering solutions to support better maneuverability, higher propulsive efficiency, and optimal operation in a broad range of environments, from coastal zones to the open ocean \cite{fish2014evolution}. Nature-inspired robotic solutions for engineering problems have been proposed by many, and examples include tuna, scallops, and dolphins \cite{vogel2000cats, Scaradozzi2017, Lauder2015,zhu2019tuna,Robertson2019,Wu2019}. However, even though marine organisms can efficiently swim in low-to-high Reynolds number (Re) regimes, engineering solutions in the moderate realm (in which both viscosity and inertia play a role in the hydrodynamics) are still lacking.

Krill, in particular, is an important model organism functioning at intermediate Re. They are hypothesized to be ecosystem engineers by inducing large-scale biogenic transport as a result of their diel vertical migrations (DVM)  of up to 1 km from the sea surface \cite{wilhelmus2014observations, bianchi_global_2016, houghton2018vertically, houghton2019alleviation, bianchi_diel_2013}. Furthermore, from an engineering standpoint, their ability to form aggregations in a wide range of configurations and migrate in coastal and open ocean regions highlights the potential for creating new compact UUV designs able to form swarms and operate in a wide range of environments \cite{Zhou2004}. 

Unlike high-Reynolds swimmers, krill locomote via drag-based metachronal swimming, named after the tail-to-head traveling wave of their appendages, operating at a phase lag \cite{vogel2020life}. In this swimming gait, the profile area of the appendages increases during the power stroke to maximize thrust. During the return stroke, the profile area decreases, reducing the drag on the appendages, such that the net force on each appendage acts in the swimming direction, creating a net thrust force sufficient to overcome the drag on the body \cite{murphy_metachronal_2011, Swadling2005}. Although there is not yet a unifying theory of fluid dynamics and force distribution for drag-based metachronal swimming, studies have shown that it is more effective than lift-based propulsion for accelerating, braking, and turning at low speeds as it can generate significant thrust over short periods, making this propulsion mechanism more adequate for maneuvering at intermediate Re \cite{byron2021,vogel2003comparative,walker_mechanical_2000,vogel2020life}.

Laboratory studies of krill via particle image velocimetry (PIV) have been invaluable to understand the effect of fluid-structure interactions on the far-field flow \cite{murphy_metachronal_2011,Murphy2013}. Murphy et al. characterized the swimming kinematics of live krill and attributed the success of the propulsion system to appendage morphology, stroke kinematics, and the resulting hydrodynamic effect from the first two \cite{murphy_metachronal_2011}. The formation of tip vortices on the pleopods, observed by Murphy in krill and Garayev and Murphy in mantis shrimp, was shown by Kim and Gharib and DeVoria and Ringuette to be an important feature in the production of thrust \cite{Murphy2013,kim_characteristics_2011,devoria_vortex_2012,Garayev2021}. Using idealized pleopod shapes, Kim and Gharib showed that the area enclosed by the tip vortex plays an important role in thrust generation \cite{Murphy2013,kim_characteristics_2011}. However, difficulty in characterizing the near-field flow has hindered efforts to fully understand thrust production in krill.

The development of simplified robotic models, as well as numerical simulations, has complemented these efforts, shedding light on the role of varying Re, phase lag, and appendage spacing on the hydrodynamics \cite{ford_hydrodynamics_2019,lim_kinematics_2009, Garayev2021, Granzier-Nakajima2020, hayashi_metachronal_2020, Alben2010}. Ford et al. looked at the effect of varying the Re on the individual jets produced by the pleopods and found that at low Re the individual jets do not interact due to viscous dissipation, but at Re around 800 they form a near-steady jet. This resulted in more significant vertical and horizontal momentum, allowing krill to generate both lift and thrust forces, necessary for locomotion and hovering \cite{ford_hydrodynamics_2019}. Phase lag, another factor contributing to metachronal swimming efficiency, has been shown to yield near-maximal efficiency and thrust production in krill swimming ranges and leads to a higher average body velocity than synchronous rhythms \cite{zhang_neural_2014,ford_hydrodynamics_2019,hayashi_metachronal_2020,Alben2010}. Appendage spacing is also a critical morphological factor affecting stroke kinematics, and studies have shown that spacing of less than 1 G/L (ratio of appendage spacing to appendage length) results in greater swimming performance during metachronal swimming \cite{Ford2021}.

Adding complexity, however, is necessary to further our understanding of the effects of morphological and kinematic characteristics on the near- and far-field flow, which is needed to establish the relationship between locomotor kinematics and swimming efficiency. Among many morphological traits of krill, the shape and flexibility of the propulsors, together with the induced fluid-structure hydrodynamic interactions in the vicinity of setae, have the potential to contribute the most to thrust and swimming efficiency. Flexibility was shown to be an important characteristic of propulsors by Kim and Gharib, who compared the thrust generation in rigid and flexible plates and found that flexible plates smooth out thrust peaks and can generate nearly constant thrust during the power stroke \cite{kim_characteristics_2011}. Setae are also important features of the pleopods of krill as they can either induce a paddle- or sieve-like effect depending on the setae density and the Re \cite{cheer_paddles_1987}. Although these traits have been investigated using simplified models, it is important to adopt an integrative framework allowing for a comprehensive understanding of the influence of different morphological features on thrust production and vortex generation.

Motivated by a robotics-inspired biology approach to address this problem, we developed a drag-based metachronal robotic platform reproducing the swimming kinematics of krill \cite{gravish2018robotics}. Our unique design implements active control of both the proximal and distal appendages and passive control of the out-of-plane motion between the two rami of the distal appendages. The assembly of 3D-printed modules allows the implementation of representative morphological features in a robotic platform able to swim in the same dynamic ranges as live krill. Additionally, due to the modular design, each appendage can be controlled independently to reproduce different kinematics for different organisms, allowing us to investigate metachronal, drag-based swimming across taxa. Going forward, we plan to use \emph{RoboKrill} to study vortex formation processes that could contribute to thrust and analyze its effect on the near-field flow structure. Our long-term goal is to engineer a new generation of AUVs that can operate in complex marine environments leveraging the swimming characteristics of metachronal swimmers. 

This paper is organized as follows. Section 2 describes the design of the driving mechanism of the robotic platform. Section 3 presents our methodology, including the robot manufacturing and the experimental setup for validation. Section 4 presents results from the kinematics analysis comparing our robotic platform and live krill. Finally, Section 5 includes a discussion on the use of \emph{RoboKrill} in laboratory experiments to further our understanding of metachronal swimming and assess the ecological role of krill.



\section{Robotic design}

Krill grow up to approximately 5 cm in length. They swim by beating their five pairs of appendages (P1 through P5 in Figure 1) at a phase lag, starting with the one at the posterior of the body (P5) \cite{Kils1981}. The appendages are biramous, paddle-shaped pleopods, each consisting of a stalk (protopodite) and two distally attached rami (exopodite and endopodite) that bear a fringe of setae (hair-like structures) (Figure 1b) \cite{lim_kinematics_2009}. The appendages have different lengths, with P2 being the longest and P5 the shortest. 

\begin{figure*}
\begin{framed}
    \centering
    \includegraphics[scale=0.5]{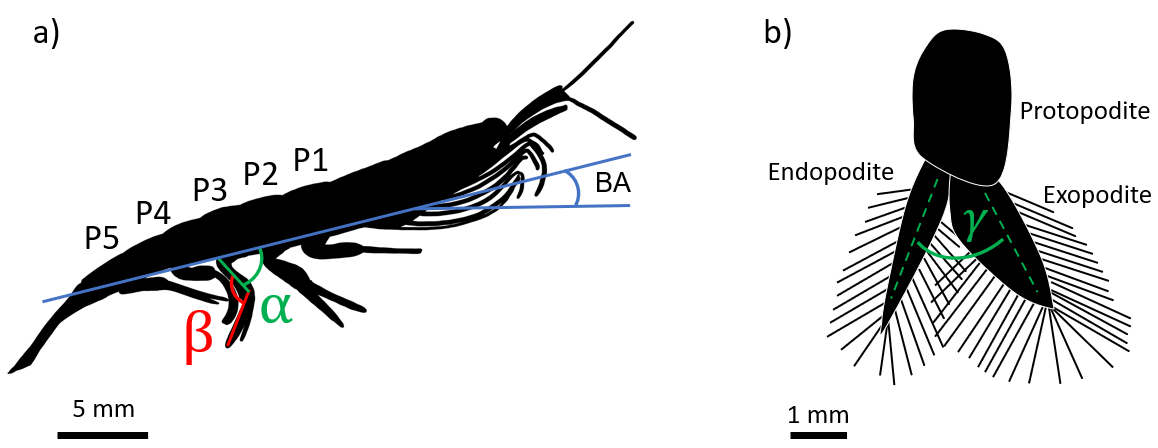}
    \caption{The morphology of \emph{Euphausia superba}. The side view of the body is shown in panel a with pleopods labeled P5 to P1. The body axis forms an angle with respect to the horizontal (BA). The body axis and the protopodite form angle $\alpha$, while $\beta$ is the angle formed between the protopodite and the distal segment (exopodite and endopodite). One of the pleopods is shown in panel b from an anterior perspective, with the exopodite extending to the right of the endopodite, forming angle $\gamma$. Both distal appendages have a fringe of setae.}
    \label{fig:my_label}
    \end{framed}
\end{figure*}

\emph{Robokrill} is dynamically scaled based on the Re. Here, this dimensionless number is defined using the velocity and length scale of the appendages and is roughly estimated to be 600 for live krill:

\begin{equation}
 Re = \frac{U_{tip}*L}{\nu} = \frac{2 \theta n L}{\nu} \;,  \tag{1}
\end{equation}

\noindent
where $L$ is the length of the pleopod, $U_{tip}$ the velocity at its tip, and $\nu$ the kinematic viscosity of the fluid. $U_{tip}$ is calculated using the stroke amplitude $\theta$ and the frequency $n$. To maintain the same Re, for the $10 \times$ scaled-up robot, the swimming frequency is reduced accordingly to dynamically match swimming krill, i.e. the beat frequency of \emph{Robokrill} is 0.57 Hz instead of the 5.7 Hz recorded for live krill \cite{murphy_metachronal_2011}. For reference, the Re of the body, based on the mean velocity and length scales of an organism, is estimated to be approximately 10,000 \cite{murphy_thesis_nodate}. It should be noted that the shape of the propulsors (exopodite and endopodite) is that of krill, drawn from high-resolution images of \emph{Euphausia superba} (San Francisco Bay Brand, Newark, CA, USA).

The stroke kinematics for \emph{E. superba} have been reported by Murphy et al. \cite{murphy_metachronal_2011}, who characterized the three major angles that describe krill motion during fast-forward swimming: $\alpha$, the angle between the body and the protopodite (proximal segment); $\beta$, the angle between the protopodite and the distal segment; and $\gamma$, the angle between the two distal segments, i.e., the endopodite and the exopodite. In fast forward swimming, $\alpha$ and $\beta$ can be characterized using sinusoidal functions. The respective function for $\alpha$ has an average peak-to-peak amplitude of 89$^{\circ}$, with average minimum and maximum values of 14$^{\circ}$ and 103$^{\circ}$, respectively. The peak-to-peak amplitude increasingly varies from 78$^{\circ}$ to a maximum of 106$^{\circ}$ from P1 to P5. The average peak-to-peak amplitude for $\beta$ is 56$^{\circ}$ with average minimum and maximum values of 105$^{\circ}$ and 162$^{\circ}$, respectively. Similarly to $\alpha$, the peak-to-peak amplitude also increases from P1 to P5 from 48$^{\circ}$ to 71$^{\circ}$. The mean beat frequency is 5.7 Hz \cite{murphy_metachronal_2011}.

The complex locomotive system that actuates each appendage in live krill is implemented in \emph{Robokrill} by a multi-link mechanism using a transmission gear box actuating both the proximal and the distal segments. The active actuation of the robotic appendages is achieved using servo motors allowing us to control the range of $\alpha$ and $\beta$ resulting in the accurate reproduction of krill kinematics. Angle $\gamma$ performs out-of-plane motion, posing challenges for active actuation and it is thus passively actuated via the hydrodynamic interaction between the structure (pleopods) and the fluid. 

The kinematic relationships governing the positions of the gears are set as \cite{erdman1997mechanism,sandor1984advanced}:%
\begin{equation}
\frac{\Delta \phi _{k+1}-\Delta \phi _{k}}{\Delta \phi _{k-1}-\Delta \phi
_{k}}=-\frac{N_{k-1}}{N_{k+1}}=-\dfrac{r_{k-1}}{r_{k+1}} \; .\tag{2}
\end{equation}
\noindent
This equation gives the relationship between the driving and driven gears, $\Delta \phi _{k+1}$ and $\Delta \phi _{k-1}$, respectively, and the link connecting both, $ \Delta \phi _{k}$ (Figure 2). Here, we consider the angle of rotation of the first link, $\phi_{0}$, such that the values $\Delta \phi _{k+1}=\phi _{k+1}-\phi _{0,k+1}$, $\Delta \phi_{k-1}=\phi _{k-1}-\phi _{0,k-1}$, $\Delta \phi _{k}=\phi _{k}-\phi _{0,k}$, capture the finite rotation of the $k^{th}$ link. Also, $N$ and $r$ are the number of teeth and radius of the corresponding gear, respectively. The negative sign accounts for the counter rotation of the gears. Meanwhile, the equation governing the position of the pleopod $\mathbf{R}_{pi}$ of a given appendage $i$ is (Figure 3,b):%
\begin{equation}
\mathbf{R}_{pi} = \mathbf{R}_{1i}+\mathbf{R}_{2i} \;, \tag{3}
\end{equation}%

\begin{figure}
\begin{framed}
    \centering
    \includegraphics[scale=0.5]{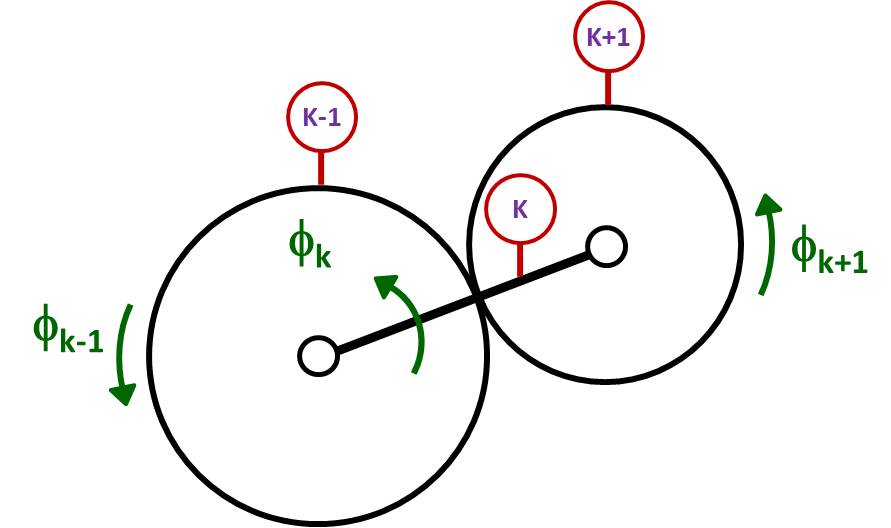}
    \caption{Epicyclic gear pair. Gear pair with input rotation $\phi$ and link number $k$. Image adapted from \cite{sandor1970kinematic}.}
    \label{fig:my_label}
    \end{framed}
\end{figure}

\noindent
where:%
\begin{equation}
\begin{tabular}{lll}
$\mathbf{R}_{1i}=\mathbf{R}\left( \theta _{1i}\right) \mathbf{r}_{1i}$ ;&   
$\mathbf{R}_{2i}=\mathbf{R}\left( \theta _{2i}\right) \mathbf{r}_{2i}$ \\  
$\mathbf{r}_{1i}=\left[ x_{1i},0\right] ^{T}$ ; &
$\mathbf{r}_{2i}=\left[ x_{2i},0\right] ^{T}$ \\ 
\end{tabular}
\tag{4}
\end{equation}%
\noindent
Here, we consider $\theta _{1i}=2\pi -\alpha _{i}$, $\theta _{2i}=\theta _{1i}-\delta
_{i}=\pi +\beta _{i}-\alpha _{i}$, and $\delta _{i}=\pi -\beta _{i}$. Also,  $\mathbf{r}_{1i}$ and $\mathbf{r}_{2i}$ represent the local reference frame vectors of the protopodite(s) and the endopodite(s), respectively (see Figure 3,a),  $\mathbf{R(\theta_{1i})}$ and $\mathbf{R(\theta_{2i})}$ are the global rotation matrices for the protopodite and endopodite, respectively.

\begin{figure*}
\begin{framed}
    \centering
    \includegraphics[scale=0.55]{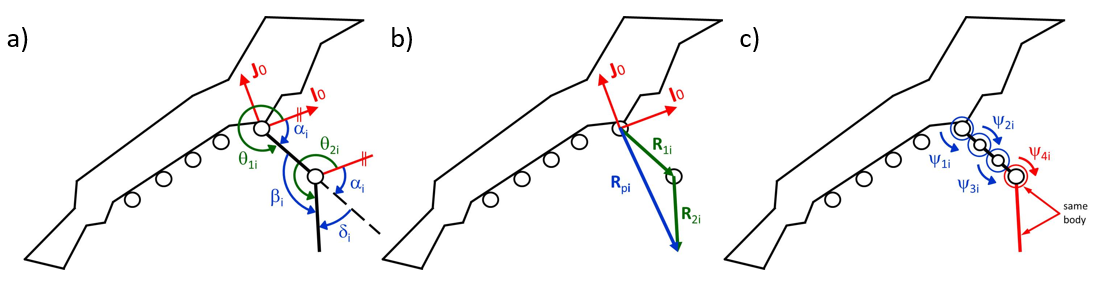}
    \caption{The locomotive system of \emph{Robokrill}. Diagrams a and b show the angles and vectors of the $i$-th appendage that are used to represent the movement of the pleopods. Diagram c shows the control of the distal appendages through a gear transmission, where $\psi$ represents the angular displacement of the gears.}
    \label{fig:my_label}
    \end{framed}
\end{figure*}

Finally, the movement of both the endopodite and exopodite is controlled by a gear transmission. The angular displacements of the gears are obtained using the relationships described in Equation 2:%
\begin{align}
\dfrac{\Delta \psi _{2i}-\Delta \theta _{1i}}{\Delta \psi _{1i}-\Delta
\theta _{1i}} = -r_{e1}  \nonumber \\
\dfrac{\Delta \psi _{3i}-\Delta \theta _{1i}}{\Delta \psi _{2i}-\Delta
\theta _{1i}} = -r_{e2}  \tag{5} \\
\dfrac{\Delta \psi _{4i}-\Delta \theta _{1i}}{\Delta \psi _{3i}-\Delta
\theta _{1i}} = -r_{e3}  \nonumber
\end{align}%

\noindent
where $\psi$ is the rotation of the gear along link 1 (endopodite), $\theta$ is the angle of the first link measured from the global reference frame (Figure 3,a), $r$ is the gear ratio, $\Delta \theta _{1i}=\theta _{1i}-\theta _{1i,0}$, $\Delta \psi _{1i}=\psi _{1i}-\psi
_{1i,0}$, $\Delta \psi _{2i}=\psi _{2i}-\psi _{2i,0}$, $\Delta \psi
_{3i}=\psi _{3i}-\psi _{3i,0}$, $\Delta \psi _{4i}=\Delta \theta _{2i}$, where $\Delta \theta
_{2i}=\theta _{2i}-\theta _{2i,0}$, and $%
r_{e1}=r_{1}/r_{2}$, $r_{e2}=r_{2}/r_{3}$, $r_{e3}=r_{3}/r_{4}$. Equations (3)
and (5) allow for the solution of direct kinematics of the mechanism by supplying the
angles $\alpha _{i}~$and $\beta _{i}$, and allows calculation of the angle $%
\psi _{1i}$, which moves the second bar (endopodite). The first bar
(protopodite) is moved by angle $\alpha _{1i}$. The metachronal trajectory at the end of pleopod 1 (Figure 4), is given by $\mathbf{R}_{p1}=\left[
x_{p1},y_{p1}\right] ^{T}$. The adopted design parameters are $x_{11}=3.2$ cm, $x_{21}=4.95$ cm and the values
for $\theta _{1i,0}, \theta _{2i,0}, \psi _{1i,0}, \psi _{2i,0},$ and $\psi _{3i,0}$
are assumed to be zero.

\begin{figure}
\begin{framed}
    \centering
    \includegraphics[scale=0.4]{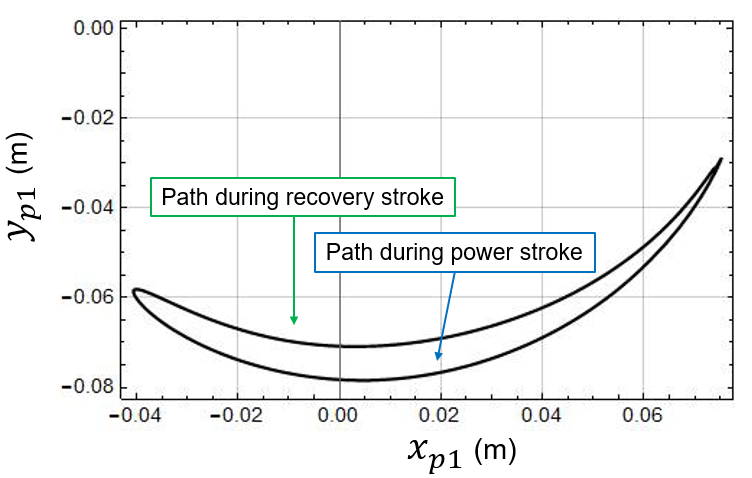}
    \caption{Trajectory of the pleopod tip. The distal appendage extends to the maximum distance away from the body during the power stroke and retracts closer to the body during the recovery stroke. The trajectory is obtained by tracking the tip of the first pleopod P1 of \emph{Euphausia superba} using a side view during fast-forward swimming (FFW).}
    \label{fig:my_label}
    \end{framed}
\end{figure}

\subsubsection{Mechanical design and manufacturing}

The architecture of the robot is composed of a set of five supports, ten sets of transmission gears, and five sets of appendage pairs (see Figure 5,a). Each appendage consists of a chain of three links, with the first link being an epicyclic gear train (Figure 5,b). Each support contains two actuators and two sets of gears, one for $\alpha $ and one for $\beta $. Both transmissions are arranged in parallel and end at axis A.

\begin{figure*}
\begin{framed}
    \centering
    \includegraphics[scale=0.5]{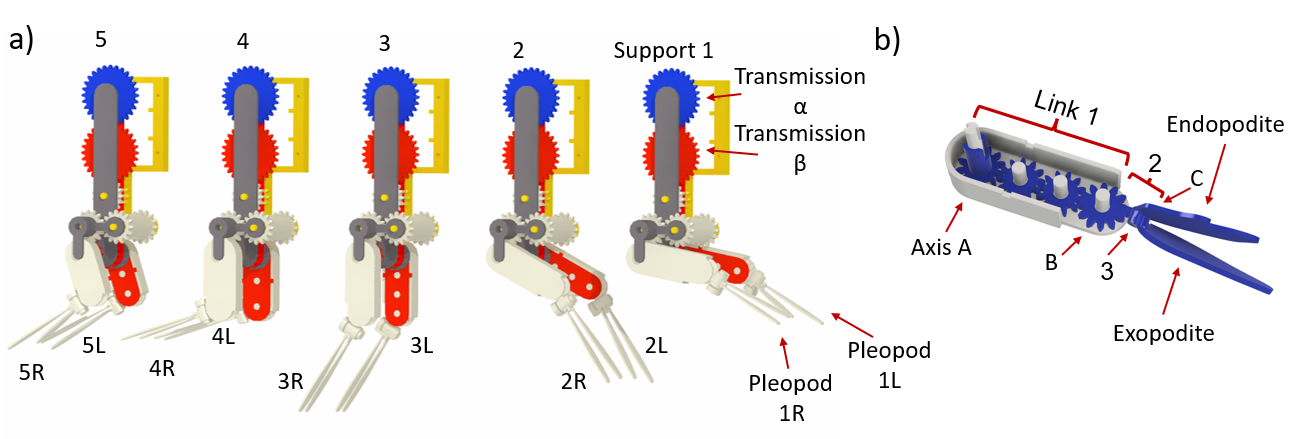}
    \caption{\emph{RoboKrill} Computer-Aided Design (CAD). Panel a shows the CAD drawing of all five pairs of appendages, labeled 1R through 5R and 1L through 5L, and the supports, labeled 1 to 5. Each support houses two servos that power an appendage pair and control $\alpha$ and $\beta$. Panel b shows the interior of a pleopod containing link 1 and 2 as well as axis A and B, that move the exopodite and endopodite. Link 3 connects the endopodite to the exopodite, which rotates along axis C. }
    \label{fig:my_label}
    \end{framed}
\end{figure*}

The first link (protopodite) of each appendage rotates $\alpha $ degrees
about axis A and contains the rotational
axis for the gears that drive angle $\beta $ and the rotational axis B for the second link (endopodite).
This link also works as the arm of the epicyclic gear system. For appendages
one through three, the epicyclic system contains four gears (with the last one being
attached to link two). Appendages four and five have an epicyclic system of three gears,
with the last one being attached to link two. This is due to the short length of protopodites P4 and P5 with respect to the rest, as per the reported measurements of live krill in \cite{murphy_metachronal_2011}. The endopodite paddle is extended at the end of the second link, while the axis of rotation for the third link is placed perpendicular to the axis of rotation B. Finally, the third link (exopodite) rotates an angle $\gamma $
about axis C with its paddle
attached.

Manufacturing was completed by 3D printing all the parts, keeping small tolerances to reduce vibrations and loss of movement of the gear train. The supports house the servos and gears, and were designed to be an above-water structure. Bearings are used to reduce friction between gears and the axes. This prevents the stagnation of the transmission, which was designed taking the servomotor speed as input with an amplification of 2.5 to achieve the desired angular speed for the links. The gears of the epicyclic system were designed by fixing the primitive radius and teeth number. The design freedom provided by additive manufacturing allowed for main consideration of the primitive radius and teeth number instead of the standard modulus of the gears. The minimum teeth number was 12, and the primitive radius $r_{pi}$ for the gears in each appendage was calculated by taking into account the length of the first link $x_{1i}$ (protopodite) and the number of gears $N_{Gi}$ contained by the link:%
\begin{equation}
r_{pi}=\frac{x_{1i}}{2N_{Gi}-2}=\frac{\text{0.032}}{2\left( 4\right) -2}=%
\frac{2}{375}m  \tag{6}
\end{equation}%

\noindent
This equation uses the length of the first link divided by the number of times the primitive radius fits between the axis of the first and the last gear. Finally, the modulus can be obtained using the following relation, where $N$ is the number of teeth:%

\begin{equation}
m=\frac{2r_{pi}}{N}  \tag{7}
\end{equation}%

\noindent
Figures 6,a and 6,b show the kinematic transmission diagram of the robot. Figure 6c shows the printed transmission with servomotors connected.

\begin{figure*}
\begin{framed}
    \centering
    \includegraphics[scale=0.5]{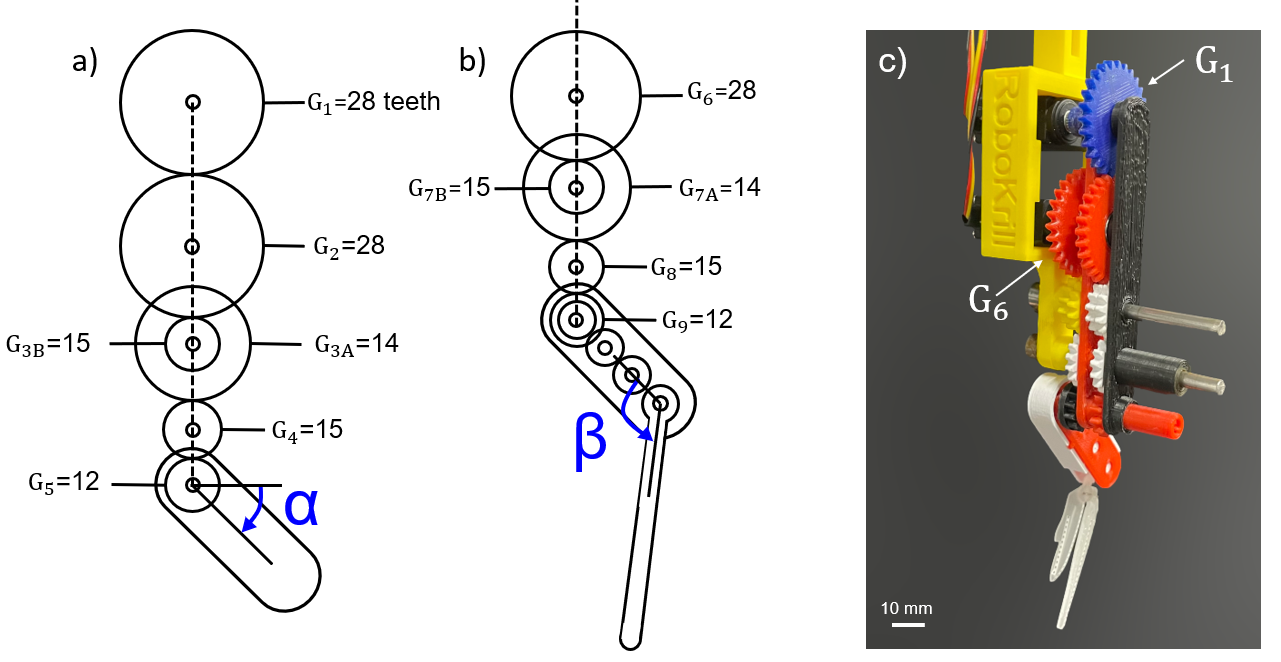}
    \caption{Transmission gear box of \emph{RoboKrill}. Panel a shows a sketch of the gear train that actuates the protopodite, forming angle $\alpha$ with the main body axis. Panel b shows the gear train that actuates the distal appendages, forming angle $\beta$ with the protopodite. The number of teeth for each gear in the train are indicated in both cases. Panel c shows \emph{RoboKrill} and the gear trains that compose the transmission of the protopodite and distal appendages.}
    \label{fig:my_label}
    \end{framed}
\end{figure*}

\subsection{Exopodite and endopodite}

As \emph{Euphausia superba} propels forward, the endopodite and exopodite abduct and adduct to effectively change the profile area of the appendages to generate thrust. This motion is characterized by $\gamma$, the angle between the exopodite and the endopodite (e.g., Figure 7 for \emph{Caridina cantonensis sp}). In \emph{Robokrill} this process is actuated on the horizontal plane, lateral to the protopodite.

According to the study by Murphy et al., exopodite abduction occurs at the beginning of the power cycle, and remains at its maximum position of 77 $^{\circ}$ during the power stroke \cite{murphy_metachronal_2011}. The exopodite and endopodite then adduct during the recovery stroke. This motion induces cupping of the appendages creating a V-shaped structure \cite{Murphy2013}, reminiscent of those observed in swimming fish that have been shown to produce greater thrust compared to flat fins (e.g., see \cite{Esposito2012}).

\begin{figure}
\begin{framed}
    \centering
    \includegraphics[scale=0.4]{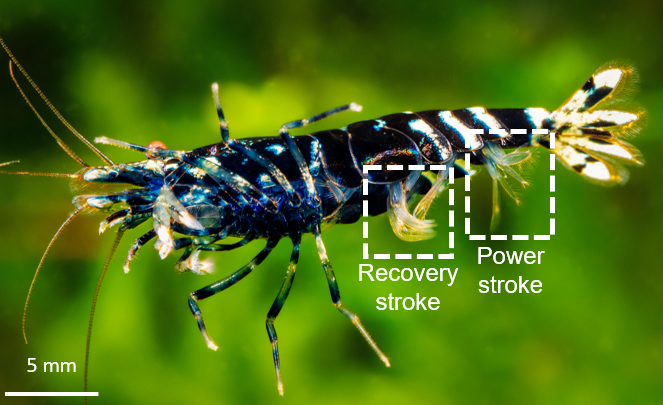}
    \caption{Drag-based metachronal swimming in \emph{Caridina cantonensis sp}. The appendages abduct during the power stroke, with both endopodite and exopodite extended, and adduct during the recovery stroke, including bending of the distal appendage. This creates an asymmetry between the power stroke and recovery stroke, creating a caudoventral jet that propels the organism forward. Image adapted from \cite{blacktiger}.}
    \label{fig:my_label}
    \end{framed}
\end{figure}

Appendage cupping forms angle $\zeta$ between the planes where the endopodite and the exopodite are actuated (see Figure 8,a). Here, photographic evidence was used to quantify it by measuring the angle between the midplanes of the endopodite and the exopodite. \emph{Robokrill} was set to match the mean quantified value of  37$^{\circ}$ (Figure 8,b). 

\begin{figure*}
\begin{framed}
    \centering
    \includegraphics[scale=0.5]{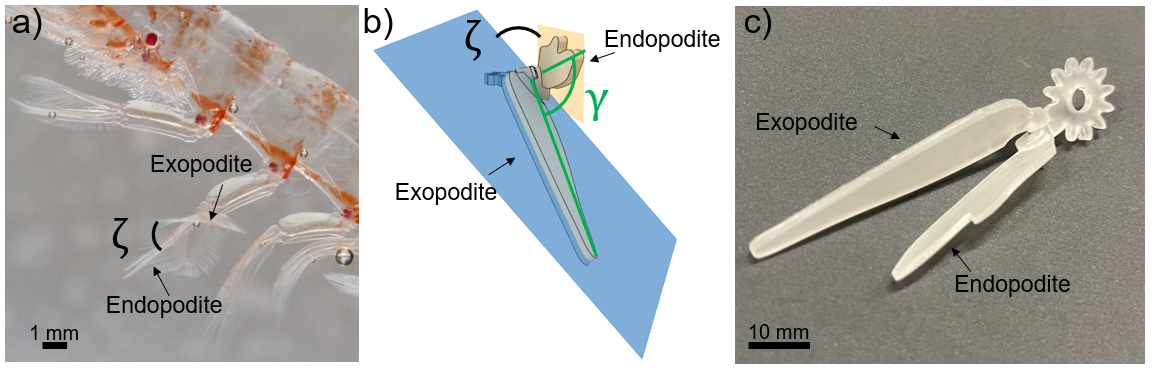}
    \caption{Cupping of the distal appendage. The angle $\zeta$ is the angle between the midplanes of the endopodite and the exopodite. Panel a shows this angle from a side view of \emph{Euphausia superba}, image adapted from \cite{superba}. Panel b shows the CAD model illustrating both the kinematic angle $\gamma$ and the fixed cupping angle $\zeta$, with the latter set to 37$^{\circ}$. Panel c shows the 3D-printed part.}
    \label{fig:my_label}
    \end{framed}
\end{figure*}

\section{Methods}

Modular CAD designs were printed using two different techniques. The transmission gears, the supports, and the protopodite housing were printed with polylactic acid (PLA) with a Prusa i3 MKS3+ 3D printer (Prusa Research, Prague, Czech Republic) for fast prototyping, while the endopodite and the exopodite were printed by stereolithography (SLA) using clear resin, for its optical properties, with a Form 3 3D printer (FormLabs, Somerville, MA, USA).

Each pair of appendages is actuated by two servos (HS-5087MH, Hitec RCD, San Diego, CA, USA), controlled by a microcontroller (ELEGOO Mega 2560, Elegoo Industries, Shenzhen, China) programmed using Simulink (MathWorks, Natick, MA, USA) via two repeating sequence interpolated blocks, one for $\alpha$ and one for $\beta$, prescribing the angles adapted from \cite{murphy_metachronal_2011} every 10 ms.  The full CAD library and assembly as well as the list of purchased components can be accessed in the open-access repository by Oliveira Santos et al. \cite{oliveira}.

The swimming kinematics of \emph{Robokrill} were analyzed and compared to those reported for live krill for validation \cite{oliveira,murphy_metachronal_2011}. One robotic appendage was tethered to a traverse beam and submerged in water at room temperature, leaving the structure housing the servos out of the water. Appendage motion was recorded at 500 fps using a scientific camera (Photron Fastcam Nova R2, Photron USA, Inc, San Diego, CA, USA, see Figure 9,a). Black markers on the surface of the robotic pleopod were tracked, both on the protopodite and the exopodite (Figure 9,b) by digitizing the video recordings via DLTdv8 for MATLAB using automatic point tracking \cite{Hedrick_2008}. The robot was allowed to run for a period of time before recording began. 


\begin{figure*}
\begin{framed}
    \centering
    \includegraphics[scale=0.65]{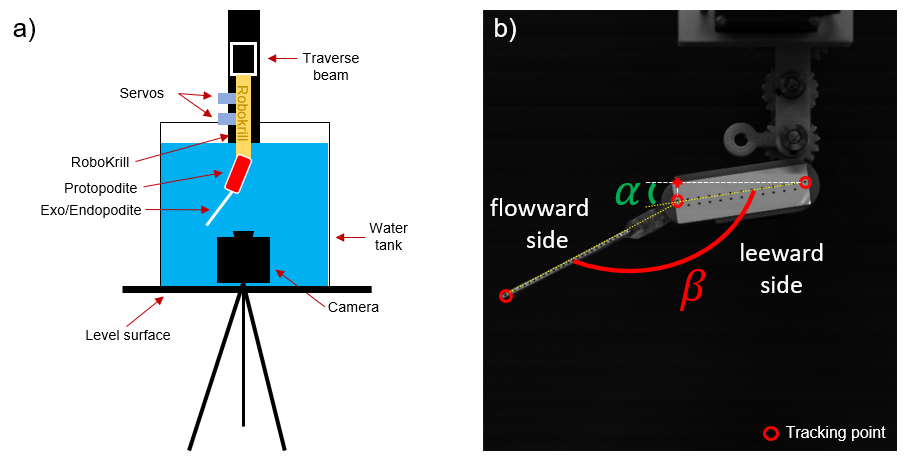}
    \caption{Experimental setup and tracking points. Panel a shows \emph{RoboKrill} tethered to a traverse beam and submerged in water. A camera is used to capture the motion of the appendages to track their movement. Panel b shows the protopodite and endopodite with tracking points (red) and the angles $\alpha$ and $\beta$.}
    \label{fig:my_label}
    \end{framed}
\end{figure*}

\section{Results}

The kinematic analysis of \emph{Robokrill} shows it can perform the prescribed movement with minimal error. By prescribing the rotation of the servos every 10 ms, the position of the robotic pleopod is set to correspond to the values reported for live krill \cite{murphy_metachronal_2011}. While friction is expected to result in some loss of movement, our measurements show that this loss is negligible. Indeed,  $\alpha$ evolves very closely to the reported values of live krill (Figure 10). The discrepancy observed in the evolution of $\beta$ can be attributed to the tolerance error between gears, which is directly proportional to the number of gears in the train. Since $\alpha$ has the smallest gear train, the path prescribed is executed without large errors. However, $\beta$ has a longer gear train and accumulates more errors due to tolerances. In addition, since smaller tolerances require more torque from the servos, tolerances for $\beta$ are greater than for $\alpha$ due to the long gear train as well as the change of plane of motion.



Direct comparison of the kinematic data obtained from \emph{RoboKrill} to measurements of live krill, shows a good fit for $\alpha$, with a maximum difference from live krill data of 6.8$^{\circ}$, occurring at the beginning of the cycle, and a mean difference over one cycle of 2.1$^{\circ}$, corresponding to 2.8\% error. The peak-to-peak amplitude of $\alpha$ for \emph{RoboKrill} is 77$^{\circ}$ and the peak-to-peak amplitude for P1 in live krill is 78.4$^{\circ}$. Amplitude for $\alpha$ of \emph{RoboKrill} can be of up to 180$^{\circ}$, but comparing programmed amplitude and actual amplitude gives us a sense of loss of movement due to friction of the gears. This loss of movement represents 1.4$^{\circ}$ in this case.

Measurements of the angle evolution between the protopodite and the two distal segments display slight deviations from the programmed path (Figure 10). This is attributed to the tolerance spacing between gears, which causes a delay in the transmission of the movement, especially during transitions between strokes. The effect can be seen at the beginning of the plotted $\beta$ cycle, during stroke reversal (from power to recovery), and again at the end of the cycle. The bump between 1.25 s and 1.5 s is due to the abrupt stop of angle $\alpha$, which reverses at the end of the power cycle, influencing the movement of $\beta$. During the power and return strokes, $\beta$ shows a good fit with experimental data. The average difference is 3.5$^{\circ}$ compared to live krill, an average error of 8\%, where the maximum amplitude of live krill is 47.5$^{\circ}$ and the maximum amplitude of \emph{RoboKrill} is 44.4$^{\circ}$, a 3.1$^{\circ}$ difference. 

\begin{figure*}
\begin{framed}
    \centering
    \includegraphics[scale=0.6]{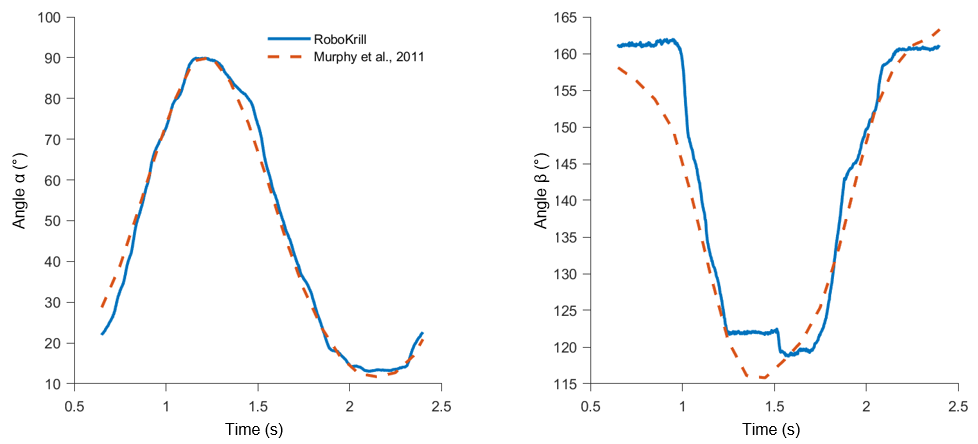}
    \caption{Evolution of the kinematic angles $\alpha$ and $\beta$ for one swimming cycle. Data from \cite{murphy_metachronal_2011} has been adapted to match the frequency of \emph{RoboKrill}, which is 1/10$^{th}$ the frequency of live krill. The angle between the body axis and the protopodite, $\alpha$, has a mean error of 2.8\%, whereas $\beta$, the angle between the protopodite and the distal segment, has a mean error of 8\%.}
    \label{fig:my_label}
    \end{framed}
\end{figure*}

While the evolution of $\gamma$ has not been quantified for pleopod 1 (P1), data from \cite{murphy_metachronal_2011} for P2 show that $\gamma$ varies between 0$^{\circ}$ and 77$^{\circ}$, with abduction and adduction coinciding with the start of the power stroke and the return stroke, respectively. In \emph{Robokrill}, this variation in $\gamma$ is obtained via passive actuation of the pleopod and thus slightly varies between cycles (Figure 11). 


\begin{figure*}
\begin{framed}
    \centering
    \includegraphics[scale=0.75]{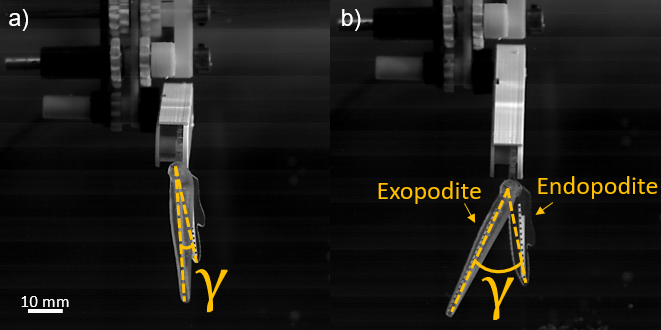}
    \caption{Actuation of $\gamma$. Panel a shows $\gamma$ with a small amplitude, at the end of the return stroke. The appendages adduct at the beginning of the return stroke (not shown) and remain adducted until the start of a new power stroke. Panel b shows the abduction of the exopodite during the power stroke. Abduction occurs at the beginning of the power stroke (not shown) and remains until the beginning of the return stroke.}
    \label{fig:my_label}
    \end{framed}
\end{figure*}

\section{Summary and Closing Remarks}

In this paper, we present \emph{RoboKrill}, a robotic platform for the study of metachronal swimming at intermediate Re that uses \emph{Euphausia superba} as a model organism. \emph{RoboKrill} was developed to be a modular 3D-printed robot that incorporates the kinematic and morphological characteristics of krill and can be used to inform on the importance of several aspects of metachronal, drag-based swimming, including the near-field hydrodynamics and formation of vortices around the pleopods.

Kinematic analysis of \emph{RoboKrill} shows it can successfully reproduce the kinematics of live krill during fast-forward swimming. \emph{RoboKrill} actively reproduces the motion of $\alpha$, the angle between the body axis and the protopodite accurately, with an average error of 2.8\% and the motion of $\beta$, the angle between the protopodite and the distal appendage, with an average error of 8\%. The out-of-plane motion resulting in the angle $\gamma$ is also actuated, albeit passively, resulting in a variation of the angle between the endopodite and exopodite of approximately 37$^{\circ}$, with abduction occurring at the beginning of the power stroke and adduction at the beginning of the recovery stroke. Additionally, the angle $\zeta$ of cupping of the endopodite with the exopodite was measured through photographic evidence to be 37$^{\circ}$ and is also represented. 

Comparison to the kinematic characterization of self-propelled organisms highlights the potential of \emph{Robokrill}, a dynamically similar 10$\times$ scaled-up model, to address relevant biological questions. Previous robotic models used in the study of metachronal swimming have been invaluable in providing information on beat frequency, phase lag, and appendage spacing, but simplification of morphological features limits their scope in studying vortex generation and near-field dynamics. \emph{RoboKrill} will complement existing knowledge on metachronal swimming by allowing us to isolate morphological features such as the propulsor shape and its flexibility as well as the presence of setae to understand individual contributions of each of these features to the overall swimming performance.

Going forward, we plan to use \emph{RoboKrill} to investigate the role of krill in the transport of oxygen, carbon, and inorganic matter during diel vertical migrations. We also intend to study the effects of varying the temperature of the water and thus the Re to understand how krill propulsive efficiency changes seasonally. This will allow us to assess the role of mesozooplankton as ecosystem engineers and the effects of increasing sea surface temperature on the locomotion of metachronal swimmers.

Manufacturing of the presented metachronal drag-based platform is accessible to both institutions and individuals who want to experiment \cite{oliveira}. Its modular design allows for the quick prototyping of different appendage shapes and sizes. Furthermore, the control of both its proximal and distal segments facilitates the analysis of different swimming kinematics, such as those corresponding to the species \emph{Euphausia pacifica sp} and \emph{Odontodactylus scyllarus sp}, also known as peacock mantis shrimp, and larger organisms such as \emph{Homarus americanus sp}, commonly known as lobster. In short, \emph{Robokrill} can be employed to obtain a comprehensive characterization of metachronal, drag-based swimming and create a unifying theory for this locomotion mechanism.

\section{Acknowledgements}

We gratefully acknowledge Valentina Di Santo, Yunxing Su, and Nils Tack for insightful discussions that strengthened the foundation of this work. We also thank Yair Sanchez Juarez and Pedro Enrique Ávila Hernández for their work on the preliminary versions of \emph{RoboKrill}. Funding for this research was provided by the University of California Institute for Mexico and the United States (UC MEXUS) and the Consejo Nacional de Ciencia y Tecnologia (CONACYT) through the UC MEXUS-CONACYT collaborative grants program (UCMEXUS CN-18-138).

\bibliographystyle{acm.bst}
\bibliography{references.bib}

\end{document}